%% file: paper.tex
\documentclass[]{bytedance_seed}

\usepackage[toc,page,header]{appendix}

\usepackage{minitoc}
\usepackage{microtype}      
\usepackage{xcolor}         

\usepackage{algorithm}
\usepackage{algpseudocode}
\usepackage{graphicx}
\usepackage{amsmath}
\usepackage{amssymb}
\usepackage{multicol}
\usepackage{float}
\usepackage{array}
\usepackage{booktabs}
\usepackage{framed}
\usepackage{mdframed}
\usepackage{tikz} 
\usepackage{amsmath}
\usepackage{multicol,multirow}
\usepackage{setspace}
\usepackage{makecell}
\usepackage{xcolor}
\usepackage{amsmath}
\usepackage{caption}

\title{ Human-assisted Robotic Policy Refinement via 

Action Preference Optimization }

\author[1,2,3,4,*]{Wenke Xia}
\author[2]{Yichu Yang}
\author[2]{Hongtao Wu}
\author[2]{Xiao Ma}
\author[2]{Tao Kong}
\author[1,3,4, \dagger]{Di Hu}

\affiliation[1]{Gaoling School of Artificial Intelligence, Renmin University of China, Beijing}
\affiliation[2]{ByteDance Seed}
\affiliation[3]{Engineering Research Center of Next-Generation Intelligent Search and Recommendation, MOE}
\affiliation[4]{Beijing Key Laboratory of Research on Large Models and Intelligent Governance}

\contribution[*]{Work done at ByteDance Seed}
\contribution[\dagger]{Corresponding authors}

\abstract{
    Establishing a reliable and iteratively refined robotic system is essential for deploying real-world applications. 
    While Vision-Language-Action (VLA) models are widely recognized as the foundation model for such robotic deployment, their reliance on offline expert demonstrations critically limits their capacity for \textit{post-deployment refinement}. 
    To mitigate this limitation, we introduce \textbf{Action Preference Optimization (APO)}, a method designed to refine VLA models by human-assisted preference alignment gathered through interaction with environments.
    This method begins with a human-robot collaboration framework for reliable failure correction and interaction trajectory collection through human intervention.  
    However, directly leveraging these interaction trajectories for preference optimization is non-trivial due to the challenges of irreversible robotic actions and token distribution mismatch. To solve this, APO proposes an adaptive reweighting algorithm with binary desirability signals derived from interaction, empowering VLA models effectively suppress failure-prone actions while enhancing corrective action adaptation.
    Ultimately, APO equips VLA models with the crucial capability to learn from failure, paving the way for their iterative refinement and reliable deployment in dynamic environments.
    The experiments conducted in simulation and real-world scenarios prove superior generalization and robustness of our human-assisted framework across a variety of manipulation tasks. We believe this work could bring insights for efficient and stable optimization of VLA models through human-robot collaboration. 
}

\date{\today}
\correspondence{Di Hu at \email{dihu@ruc.edu.com}}

\checkdata[Project Page]{\url{https://gewu-lab.github.io/action_preference_optimization/}}
\checkdata[Code]{\url{https://github.com/GeWu-Lab/Action-Preference-Optimization}}

\begin{document}
\maketitle


\input{sections/introduction}

\input{sections/relatedwork}

\input{sections/approach}

\input{sections/experiments}

\clearpage

\bibliographystyle{plainnat}
\bibliography{main}

\clearpage

\beginappendix

\input{sections/appendix}

\end{document}

%% file: sections/introduction.tex
\section{Introduction}

Fostering continuous improvement is crucial for the development of robust robotic manipulation systems in real-world scenarios~\cite{cui2023no,mandlekar2020human,Xia_2025_CVPR}. Benefiting from the capacity for generalizable reasoning and scalable learning, Vision-Language-Action (VLA) models~\cite{bjorck2025gr00t,black2024pi0,cheang2024gr,kim2024openvla,wen2025dexvla,MIRwork,MIRsurvey} have been widely recognized as the foundation model for such robotic deployment systems. However, prevailing training paradigm for these models hinges on large-scale datasets of expert demonstrations. This severely limits their \textit{post-deployment refinement}, as they lack intrinsic ability to continually learn from failures or adapt to novel scenarios encountered in real world.





To enhance the continuous learning ability of robotic systems, interactive imitation learning frameworks~\cite{celemin2022interactive,kelly2019hg} have been developed to refine error-prone trajectories via iterative human-in-the-loop correction feedback. Among them, behavior cloning~\cite{hoque2023fleet,liu2022robot,ross2011reduction} has been widely utilized to fine-tune a base policy model with manually corrected intervention data in a supervised learning fashion. In contrast, recent methods~\cite{li2022efficient,luo2023rlif} seek to propose effective off-policy reinforcement learning algorithms from sub-optimal human intervention trajectories. 
However, behavior cloning fails to fully exploit failure trajectories, which are valuable signals for learning robust policies. At the same time, reinforcement learning methods encounter significant scalability limitations in training large-scale VLA models, due to the inherent instability and challenge of developing generalizable value functions.
To date, the effective adaptation of VLA models for downstream manipulation tasks remains understudied, particularly within sub-optimal human intervention paradigms.

To bridge this gap, we propose \textbf{Action Preference Optimization (APO)}, a new paradigm moves beyond the limitations of both behavior cloning and reinforcement learning. By learning from action-level preferences captured during interactions, our approach fully \textit{exploits the valuable information in failure trajectories while maintaining the optimization stability} required for large-scale VLA models.

\begin{figure*}[t]
    \centering
    \includegraphics[width=\linewidth]{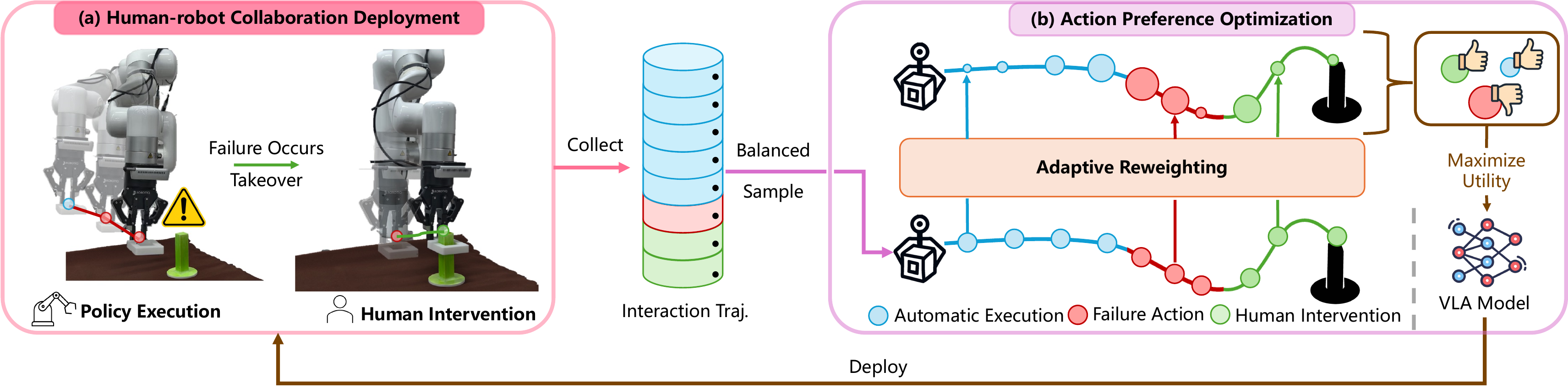}
    \caption{Our method consists of two key components: (a) the human-robot collaboration deployment framework for reliable deployment and interaction trajectory collection with human intervention. (b) the action preference optimization process with adaptive reweighting for VLA models learning from sub-optimal interaction trajectories. The size of each circle represents its weight during training.} 
    \label{fig:teaser}

\end{figure*}

Our method is founded on a human-robot collaboration framework designed to ensure reliable deployment while simultaneously generating data for policy refinement, as illustrated in Figure~\ref{fig:teaser}(a). When the robot encounters challenging situations, real-time human interventions not only guarantee successful task completion but also provide corrective trajectories. These trajectories are collected as preference pairs to refine the policy. Furthermore, to address the imbalanced distribution of action types in the collected data, we employ a balanced sampling method. This ensures a proportional representation of all interaction data for the subsequent VLA preference optimization

However, directly applying this preference data to fine-tune autoregressive VLA models presents two significant, interrelated challenges: (1) Irreversible interaction: While LLMs often require paired preference data, the irreversible nature of physical interaction makes it extremely difficult to gather perfect positive-negative action samples under identical conditions. (2) Token probability mismatch: Autoregressive VLA models discretize continuous actions into tokens, causing a fundamental mismatch between token probabilities and the true action loss, which complicates preference alignment. To address these pressing problems, we first employ Kahneman \& Tversky’s prospect theory~\cite{ethayarajh2024kto,tversky1992advances} to formulate a novel preference alignment objective that learns from binary desirability signals derived from interaction. This objective relaxes the stringent demands of preference pairs, making it suitable for learning from irreversible robotic interaction trajectories. Furthermore, we propose an adaptive reweighting method that leverages decoded continuous actions to effectively guide preference optimization in the discrete action token space. This approach addresses the challenge of action token probability mismatch via the dynamic modulation of sample-wise training weighting, thereby concentrating gradient optimization on failure-prone interaction actions. Through this targeted weight refinement, we successfully apply preference alignment optimization to VLA models, enhancing performance when \textit{learning from sub-optimal manipulation correction trajectories}.



To systematically evaluate the effectiveness of our proposed system, we conduct a comprehensive set of experiments in  RoboMimic~\cite{robomimic2021} simulation environments. 
The empirical results demonstrate that APO facilitates rapid adaptation in in-distribution scenarios while maintaining robust performance across a variety of unseen perturbations.
Furthermore, lifelong learning experiments demonstrate the framework's capacity for iterative improvement through human intervention. To evaluate the practical viability of the proposed framework, we conducted real-world experiments on fine-grained insertion tasks under a range of disruption conditions, demonstrating its robustness and applicability in real-world robotic manipulation scenarios.


%% file: sections/relatedwork.tex
\section{Related Works}\label{sec:related} 
\subsection{Vision-Language-Action Models} 
Achieving generalizable robotic manipulation remains a significant challenge within the field of robotics. Motivated by recent advances in foundation models~\cite{li2024llava,touvron2023llama,wei2022chain,pang2024,xia2024kinematic,zeng2024learning}, some works~\cite{bu2025agibot,fang2023rh20t,vuong2023open} attempt to construct large-scale real-world robotic datasets to facilitate development of generalizable Vision-Language Action (VLA) models. Building upon these datasets, recent research \cite{brohan2023rt,kim2024openvla,pertsch2025fast} formulates robotic action prediction as a next token prediction problem within the framework of VLMs. In contrast, alternative studies \cite{liu2024rdt,team2024octo} investigate the applicability of diffusion-based methods to model multi-modal action distributions, thereby facilitating robustness in manipulation tasks. While these works focus on behavior cloning from expert demonstrations, Grape~\cite{zhang2024grape} proposes a trajectory-level preference alignment method to boost generalizability by incorporating both successful and failed trials.
However, the requirements of paired trajectories under the same conditions make it infeasible in real-world scenarios. In this work, we propose the action preference optimization method to continuously refine VLA models by integrating human-in-the-loop intervention preference data.




\subsection{Preference Alignment of Large Language Models}
Contemporary methods~\cite{christiano2017deep,nakano2021webgpt,ouyang2022training,ziegler2019fine} implement Reinforcement Learning from Human Feedback (RLHF) through a two-stage method,
which first trains a reward estimation model and optimizes LLMs to maximize the given estimated reward with a reinforcement learning method~\cite{gao2023scaling,schulman2017proximal,lu2024koi}. However, this paradigm is slow and unstable in practice. DPO~\cite{rafailov2023direct} proposes a single-step alternative that reparameterizes the RLHF objective into a closed-form loss function to directly maximize the log-likelihood margin between preferred and dispreferred outputs. Extending this framework, KTO~\cite{ethayarajh2024kto} introduces the human-aware losses for learning from a binary signal of whether an output is desirable, which bypasses the need for intricate preference annotation altogether. In this work, we adapt the preference alignment optimization method for Vision-Language-Action models. Through an adaptive reweighting approach, we mitigate the irreversible interactions and token probability mismatch challenges when transferring preference learning methods from LLMs to VLA models.


\subsection{Human-robot Interactive Learning}
Interactive imitation learning~\cite{survey09,celemin2022interactive} has been proposed to refine robot actions through human feedback. While prior research~\cite{kelly2019hg,liu2022robot,ross2011reduction} necessitates constant human supervision to intervene in the robot's actions, more recent studies~\cite{gokmen2023asking,liu2024model,yel2019fast} have introduced dynamic models for automatic failure detection and real-time monitoring. In contrast, RLIF~\cite{luo2023rlif} leverages human intervention signals as rewards for off-policy RL, while HIL-SERL~\cite{luo2024precise} presents a human-in-the-loop, vision-based RL system tailored for dexterous manipulation tasks. However, these RL approaches encounter difficulties in large-scale VLA model training, primarily due to unstable gradient optimization. In this work, we propose the action preference optimization method to ensure the stable optimization of policies from action-level preferences captured during interaction.

%% file: sections/approach.tex
\section{Method}

In this section, we introduce Action Preference Optimization (APO), a method designed to facilitate continuous iterative improvement of Vision-Language-Action (VLA) models. As detailed in Algorithm~\ref{alg:human_robot}, APO aligns the model with human preferences gathered through human-robot collaboration deployment within environment.



\subsection{Human-robot Collaboration Deployment}

To ensure reliable deployment and interaction trajectory collection, our method is founded on the human-robot collaboration framework for real-time intervention and interaction data acquisition.

We first collect an expert demonstration dataset $\mathcal{D}_e = \{\tau_e^i\}_{i=1}^{i=N}$, where each trajectory $\tau_e^i$ consists of observation-action pairs with expert annotations: $\tau_e^i = \{(o_t^i,a_t^i,c_t^i)\}_{t=1}^{t=T}$, where $c_t^i = 1$ indicates that $a_t^i$ is executed by human expert. We employ behavior cloning to fine-tune the pretrained VLA model on these expert demonstrations, obtaining an initial base policy $\pi_\theta^0$. This policy is then deployed for interaction trajectory collection.

During policy execution, the human operator monitors policy execution and intervenes when the policy encounters challenging scenarios. Through this process, we could collect a set of interaction trajectories $\mathcal{D}_h = \{\tau^{i}_{h}\}_{i=i}^{i=M}$, 
where $c_t^i = 2$ represents the action is corrected by human intervention while $c^i_t = 1$ denotes the action is executed by policy. Further, we re-label the interaction trajectories to categorize the actions taken in the $K$ steps preceding human interventions as undesirable, annotated with $c_t^i=0$. 
For each trajectory, we discretize the continuous action $a$ into discrete action token $\hat{a}$.
Finally, we combine the expert demonstrations $\mathcal{D}_e$ and the interaction dataset $\mathcal{D}_h$ for further robotic action preference optimization.

\begin{algorithm}[t]
\caption{Action Preference Optimization}
\label{alg:human_robot}
\begin{algorithmic}[1]
\State \textbf{Notations:}
\State $\mathcal{D}_e$: expert demonstrations, $\mathcal{D}_h$: interaction dataset, $\pi$: interaction policy

\State \textbf{Warm-start phase}
\State Collect $\mathcal{D}_e \gets \{\tau_1^e, \dots, \tau_N^e\}$
\State Initialize BC policy $\pi^0_\theta$
\State $\theta^* \gets \arg\max_\theta \mathbb{E}_{(o, a) \sim \mathcal{D}^e} \left[ \log \pi^0_\theta(a | o) \right]$
\State $\mathcal{D}_h^0 \xleftarrow{} \mathcal{D}_e$
\State \textbf{Deployment-optimization loop}
\For{$i \gets 0$ \textbf{to} $X$}

        $\pi_{ref} \gets  \pi_{\theta}^{i}$
        
      $D_h^{i+1} \gets \textsc{Deployment}(\pi^{i}_\theta, \mathcal{D}_h^i)$
      
      $\pi^{i+1}_\theta \gets \textsc{Optimization}(\pi_{\theta}^i,\pi_{ref},\mathcal{D}_h^{i+1})$
\EndFor
\vspace{0.5em}
\Statex\hrule 
\vspace{0.5em}
\begin{minipage}[t]{0.48\textwidth}
\Function{Deployment}{$\pi_\theta, \mathcal{D}_h$}\label{algo:deployment}
    \For{n interaction rollouts}
    
    \While{task does not succeed}
        \If{human intervenes}
        
        \State $a_t \xleftarrow{} human, c_t \xleftarrow{} 2$

        \State $c_{t-K:t-1} \gets 0$
        \Else 

        \State $a_t \xleftarrow{}\pi_\theta(o_t), c_t\xleftarrow{} 1$
            
        \EndIf

    \State $\tau_i^h \xleftarrow{} \tau^h_i \cup (o_t,a_t,c_t)$

    \EndWhile
        \State $\mathcal{D}_h \xleftarrow{} \mathcal{D}_h \cup \tau_i^h$
    \EndFor
    
    
    \State \Return $\mathcal{D}_h$
\EndFunction

\end{minipage}
\hfill
\begin{minipage}[t]{0.48\textwidth}
\Function{Optimization}{$\pi_\theta$,$\pi_{ref}$,$\mathcal{D}_h$}\label{algo:optimization}
    \For{n gradient steps}
    
    \State Balanced Sample $(o_i,a_i,c_i)_{i=1}^{i=B} \sim \mathcal{D}_h$

    \State $l_i \xleftarrow{} | \pi_\theta(o_i) - a_i|_1 $

    \State $w_i \xleftarrow{} \frac{l_i}{\sum_{i=1}^{i=B} l_i}$

    \If{$c_i \neq 0$} 
            \begin{center}
                $\lambda_{D_i} = 1 - e^{-\beta_D*w_i}$
            \end{center}
    \Else 
            \begin{center}
                $\lambda_{U_i} = e^{-\beta_U*w_i}$
            \end{center}
    \EndIf

    \small \State $\theta^* = \arg\max_\theta \mathbb{E}[-v(o,a,\pi_\theta,\pi_{ref},\lambda_D,\lambda_U)]$
    
    \EndFor

    \State \Return $\pi_{\theta^*}$ 
\EndFunction
\end{minipage}
\end{algorithmic}
\end{algorithm}

\subsection{Action Preference Optimization}

To maximize the utility of sub-optimal interaction trajectories and ensure stable fine-tuning of the VLA model, we adopt the preference alignment optimization method to guide the model to learn from corrections and avoid failures.

Although previous Reinforcement Learning with Human Feedback (RLHF) methods~\cite{bai2022training,rafailov2023direct} have proven effective in LLM fine-tuning, there are additional challenges for the VLA models preference optimization in robotic manipulation:
\begin{itemize}
    \item The irreversible robotic manipulation process makes it challenging to acquire meaningful paired positive-negative actions under the same observational conditions.
    \item The mapping of continuous robotic actions to discrete tokens by autoregressive VLAs causes a mismatch between token probability and continuous action errors, complicating preference optimization in action token prediction.
    
\end{itemize}

To address these issues, we adopt Kahneman \& Tversky's prospect theory~\cite{tversky1992advances} for preference alignment optimization with binary desirability signals and propose an adaptive reweighting method to bridge the gap between discrete token prediction and continuous action regression.  We first estimate the reward function $r_\theta$ of our model $\pi_{\theta}$ as standard approach~\cite{ouyang2022training,rafailov2023direct,stiennon2020learning}:
\begin{equation}
    r_{\theta}(o, \hat{a}) = \log \frac{\pi_{\theta}(\hat{a}|o)}{\pi_{\text{ref}}(\hat{a}|o)},
\end{equation}
where $\hat{a}$ is the discrete action token and the reference model $\pi_{ref}$ is the base model $\pi_{\theta}^i$ at the beginning of each deployment-optimization loop shown in algorithm~\ref{alg:human_robot}. Following~\cite{ethayarajh2024kto,tversky1992advances}, we formulate the utility function $v$ as below to estimate the relative gain on the robotic data:
\begin{equation}
v(o, \hat{a}) = 
\begin{cases} 
\lambda_D \sigma\left(r_\theta(o, \hat{a}) - z_0  \right) & \text{if } \hat{a} \sim  \hat{a}_{\text{desirable}} \\
\lambda_U \sigma\left(  z_0 - r_\theta(o, \hat{a})  \right) & \text{if } \hat{a} \sim \hat{a}_{\text{undesirable}} ,
\end{cases}
\label{eq:v}
\end{equation}

where $\lambda_D$ and $\lambda_U$ are utilized for importance sampling, the $\sigma$ is the sigmoid function. To ensure that the model $\pi_\theta$ does not deviate excessively from the reference model $\pi_{ref}$, a penalty term $z_0$ is introduced. This term is defined as the KL-Divergence between $\pi_\theta$ and $\pi_{ref}$: $z_0 = KL(\pi_{\theta}||\pi_{ref})$. Incorporating $z_0$ into the optimization process guides the model to learn from preference pair data while simultaneously preserving knowledge acquired from prior models. We employ the following loss function $L$ to optimize the model $\pi_\theta$ using preference optimization with desirability signals:
\begin{equation}
    L(\pi_\theta,\pi_{ref}) =  \mathbb{E}_{x,y \sim D^h}[ - v(x,y)].
\end{equation}
By minimizing the loss function, we aim for the model $\pi_\theta$ to get higher rewards for desirable pairs while avoiding predicting undesirable actions, in comparison to the reference model $\pi_{ref}$.

However, directly applying the preference alignment optimization from LLMs to autoregressive VLA models is problematic, primarily due to the differences in their respective token definitions. While word tokens correspond to distinct subwords, action tokens necessitate a non-differentiable mapping to continuous ground truth actions. This creates a discrepancy between the token classification probabilities and the regression loss associated with the continuous robotic actions.

To bridge the gap between token classification and continuous action regression in autoregressive VLA models, we introduce an adaptive reweighting method. This approach guides the model to prioritize samples exhibiting large regression errors by first estimating the L1 loss of the continuous action $l$ for each sample, followed by batch-level normalization as detailed below:
\begin{equation}
    w_i = \frac{l_i}{\sum_{i=1}^{i=B} l_i}.
\end{equation}

The normalized weighting scheme operates by: 1) for desirable data, increasing the weight of samples with high action prediction errors, and 2) for undesirable data, increasing the weight of samples whose actions are proximate to the failure actions.
By adaptively adjusting the values of $\lambda_D$ and $\lambda_U$ in Equation~\ref{eq:v} using the normalized weights, we gain fine-grained control over the relative influence of each sample during training:
\begin{align}
        \lambda_D &= 1 - e^{-\beta_D * w},
\\
    \lambda_U &= e^{-\beta_U * w}.
\end{align}

By incorporating preference alignment optimization via sample-wise weight refinement, we enhance the performance and optimization stability of the VLA model, when learning from sub-optimal manipulation correction trajectories.

In conclusion, we propose the action preference optimization method as demonstrated in algorithm~\ref{alg:human_robot}. This approach leverages the human-robot collaboration deployment for reliable task execution and interaction trajectories collection, while the action preference optimization process provides stable autoregressive VLA optimization with adaptive reweighting. Through iterative human-robot collaboration deployment and action preference optimization, we could achieve continual improvement from interaction with environments for autoregressive VLA models.

%% file: sections/experiments.tex
\section{Experiments}

To comprehensively evaluate our action preference optimization method for effective downstream adaptation, we propose experiments to validate the following questions:
\begin{itemize}
    \item How effective is APO at promoting adaptation to in-distribution scenarios? Section~\ref{exp_comparison}
    \item Does APO maintain effective learning performance in novel scenarios despite various disruptions? Section~\ref{exp_generlization}
    \item Does APO demonstrate the ability to achieve iterative improvement during deployment? Section~\ref{exp_lifelong}
    \item How well does APO generalize to different VLA models? Scetion~\ref{ref:various_vla}
    \item Could APO be applied in fine-grained real-world scenarios? Section~\ref{exp_real}
    \item To what extent does the action-level preference optimization method enable APO to learn action correction? Section~\ref{analysis}
\end{itemize}

\subsection{Experiment Settings}~\label{sec:settings}
\begin{figure*}[h]
    \centering
    \includegraphics[width=\linewidth]{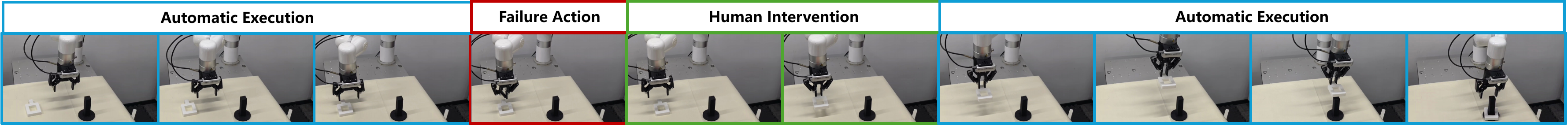}
    \caption{The demonstration of our human-assisted interaction trajectory.}\label{fig:real_world_intervention}
\vspace{-1em}

\end{figure*}

\textbf{Implementation Details.} In this work, we fine-tune the OpenVLA~\cite{vuong2023open} model for target manipulation tasks as the base model.
We employ LoRA~\cite{hu2022lora}  for parameter-efficient tuning, configuring rank $r = 32$ with a batch size of 16 across 8 NVIDIA A100 GPUs. Further, we deploy the base model to interact with environments, where human operators perform real-time corrective interventions via a SpaceMouse device to rectify failures during execution. We set $K = 10$ to identify and annotate undesirable behaviors automatically.
The human-assisted interaction trajectory is shown in Figure~\ref{fig:real_world_intervention}, which is segmented into robotic automatic execution, the failure action, and the human intervention types by the timing of human correction.
Based on the interaction trajectories collected during task execution, we fine-tune the base model $\pi_{ref}$ with our action preference optimization method, using a learning rate of $5e{\text -}5$ and a batch size of 8 across 4 NVIDIA A100 GPUs.  
To ensure the stability of preference alignment training, we employ balanced sampling to ensure that each batch contains 50\% expert actions, 25\% human intervention actions, and 25\% failure actions.

\textbf{Simulation Environments Details.}
 For a comprehensive evaluation, we validate these methods on fine-grained manipulation tasks within the RoboMimic~\cite{robomimic2021} simulation environment, such as 'make coffee' and 'toy assembly'.
In the RoboMimic environment, we fine-tune the pretrained OpenVLA model for these 4 long-horizon manipulation tasks with 300 expert demonstrations. To optimize policy with human preferences, we collect 50 trajectories per task under different seeds for RoboMimic tasks. For evaluation, we conduct 50 trials under three unseen seeds for each task, and report the average success rate.


\begin{table*}[b]
\caption{Comparison experiment results across 4 manipulation tasks in RoboMimic Simulation. The results demonstrate that our adaptive reweighting preference optimization method achieves stable improvement compared with other behavior cloning and preference optimization methods.}
\centering
\resizebox{\textwidth}{!}{
\begin{tabular}{ccccccccc}
\toprule
Methods & Coffee\_D0  & StackThree\_D0  & ThreePieceAssebly\_D0 & Square\_D0 & Mean\\ \midrule

Base policy & 44\%  & 46\% & 44\% & 28\%  & 40.5\%  \\
Dagger~\cite{ross2011reduction} & 42\% & 50\% &  36\% & 28\% & 39.0\% \\
Sirius~\cite{liu2022robot} &  34\% & 52\% & 34\%  & 38\% &  39.5\% \\
DPO~\cite{rafailov2023direct} & 52\% & 46\% & 28\%& 22\% &  37.0\% \\
TPO~\cite{zhang2024grape} & 54\% & \textbf{54\%} & 40\% &  18\% & 41.5\%\\
KTO\cite{ethayarajh2024kto} & 48\% & 52\% & \textbf{46\%}  & \textbf{32\%} & 43.5\% \\ 
\midrule
APO & \textbf{60\%} & \textbf{54\%} & \textbf{46\%} & \textbf{32\%} & \textbf{48.0\%} \\
\bottomrule
\end{tabular}}
\vspace{-1em}    
    \label{tab:robomimic_result}

\end{table*}

\subsection{Comparison Experiments}\label{exp_comparison}

We compare APO with other approaches to evaluate the effectiveness for VLA model fine-tuning. To ensure fairness, we fine-tune OpenVLA~\cite{kim2024openvla} for manipulation tasks as a base model and improve the base model with other comparison methods.
\begin{itemize}
    \item \textbf{Dagger~\cite{ross2011reduction}:} We mix the expert demonstrations with interaction trajectories, fine-tuning the base model using a behavior cloning objective.
    \item \textbf{Sirius~\cite{liu2022robot}:} we apply sample reweighting to prioritize human intervention data and fine-tune the base model using a weighted behavior cloning loss.
    \item \textbf{DPO~\cite{rafailov2023direct}:} We generate paired negative samples for interaction trajectories by perturbing the actions predicted from the base model with Gaussian noise, and fine-tune the base model using these paired data with the DPO method.
    \item \textbf{TPO~\cite{zhang2024grape}:} We select positive and negative samples in interaction trajectories based on the timing of intervention, then fine-tune model with trajectory-wise preference optimization.
    \item \textbf{KTO~\cite{ethayarajh2024kto}:} We select positive and negative samples in interaction trajectories based on the timing of intervention. Then, we sample positive and negative trajectories and optimize the base model with KTO, with the constraint $z_0 = KL(\pi_{\theta}||\pi_{ref})$.
\end{itemize}


As shown in Table~\ref{tab:robomimic_result}, we first compare the behavior cloning objective methods. The results reveal that after fine-tuning with interaction data, these methods fail to outperform the base model, which demonstrates that existing behavior cloning approaches struggle to achieve efficient adaptation in the context of large-scale VLA models. A key challenge stems from the distribution shift between expert trajectories and interaction trajectories. Without mechanisms to retain the base model's knowledge under the under standard behavior cloning objectives, this shift makes it particularly difficult for large-scale VLA models to effectively fit the complex, multimodal distribution arising from the combined expert and interaction datasets.

We further provide results of the preference optimization based methods. By integrating a regularization constraint with the reference model $\pi_{ref}$, these methods could maintain useful knowledge from the reference model while achieving improvement from interaction trajectories. 

Among all compared preference learning based methods, DPO yields the weakest performance. This result stems from its exclusive reliance on synthetic paired failure actions for optimization, which lacks exposure to real-world errors essential for teaching robots mistake avoidance through interaction. On the other side, TPO fails to deliver stable performance gains on multiple tasks while APO attains stable performance gains relative to the base model. 
The TPO method employs negative samples to regularize model preference alignment optimization, but introduces instability through random sampling. In contrast, APO utilizes KL divergence to estimate the mean margin between the updated model and the reference model, which not only enables more stable learning but also better preserves prior knowledge.
Compared with KTO, APO leverages the adaptive reweighting method to achieve more precise control over the importance weights of both positive and negative samples, delivering more notable performance improvements.

\begin{figure*}[t]
    \centering
    \includegraphics[width=\linewidth]{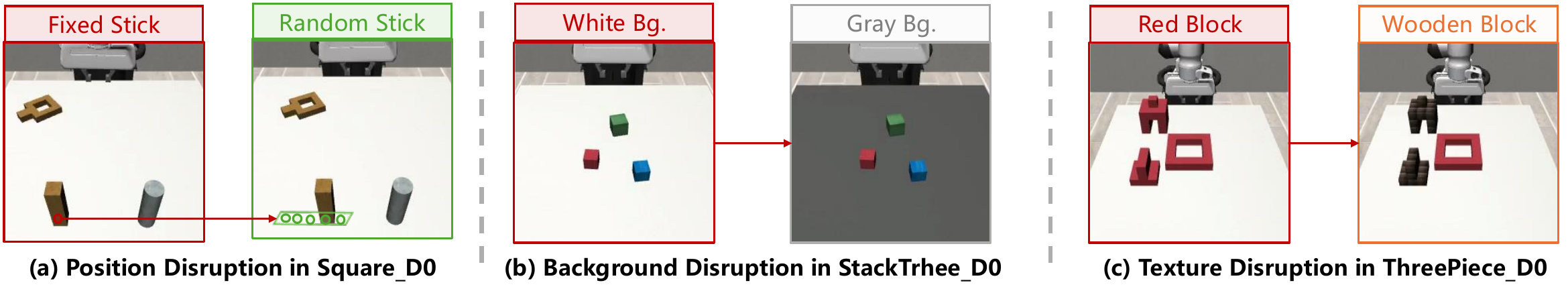}

    \caption{
    In the position disruption setting, we change the position of the stick from a fixed point \protect\tikz{\protect\filldraw[color=red, fill=white] (0,0) circle (2pt);} to a random position from the rectangle \protect\tikz{\protect\draw[color=green] (0,0) rectangle (0.15,0.15);} in the Square\_D0 task as illustrated in (a).
    In the background disruption setting, we replace the background with the \textcolor{gray}{gray one} in the StackThree\_D0 task as shown in (b). In the texture disruption setting, we replace the \textcolor{red}{red blocks} with the \textcolor{brown}{wooden ones}.
    }

    \label{fig:disruption}
    \vspace{-1em}
\end{figure*}

\begin{figure*}[t]
    \begin{minipage}[t]{0.48\textwidth} 
            \captionof{table}{The results on disruption scenarios.}
        \centering
        \scriptsize
        \begin{tabular}{ccccc}
            \toprule
            Methods   & Pos Dis.  & Bg Dis. & Tex Dis. & Mean \\ 
            \midrule
            Base policy & 12\% & 42\% & 10\% & 21.3\% \\
            Dagger & 18\% & \textbf{46\%} & 4\% & 22.7\% \\
            Sirius & 12\% & 42\% & 2\% & 18.7\% \\
            DPO & 14\% & 26\% & 2\% & 14.0\% \\
            TPO & 18\% & 32\% & 8\% & 19.3\% \\
            KTO & 20\% & \textbf{46\%} & 6\% & 24.0\% \\
            \midrule
            APO & \textbf{26\%} & \textbf{46\%} & \textbf{12\%} & \textbf{28.0\%} \\
            \bottomrule
        \end{tabular}
        \label{tab:var_result}
    \end{minipage}%
    \hfill
    \begin{minipage}[t]{0.48\textwidth} 
        \captionof{table}{The results on original tasks.} 
        \centering
        \scriptsize
        \begin{tabular}{ccccc}
            \toprule
            Methods   & Square  & StackThree & ThreePiece & Mean \\ 
            \midrule
            Base policy & 28\% & 46\% & 44\% & 39.3\% \\
            Dagger & 16\% & 46\% & 30\% & 30.7\% \\
            Sirius & 18\% & 48\% & 18\% & 28\% \\
            DPO & 20\% & 50\% & 30\% & 33.3\% \\
            TPO & 30\% & 36\% & 40\% & 35.3\% \\
            KTO & 30\% & 46\% & \textbf{42\%} & 39.3\% \\
            \midrule
            APO & \textbf{34\%} & \textbf{62\%} & 40\% & \textbf{45.3\%} \\
            \bottomrule
        \end{tabular}

        \label{tab:var_ori_results}
    \end{minipage}
    \vspace{-1em}
\end{figure*}

\begin{figure*}[t]
    \centering
    \includegraphics[width=\linewidth]{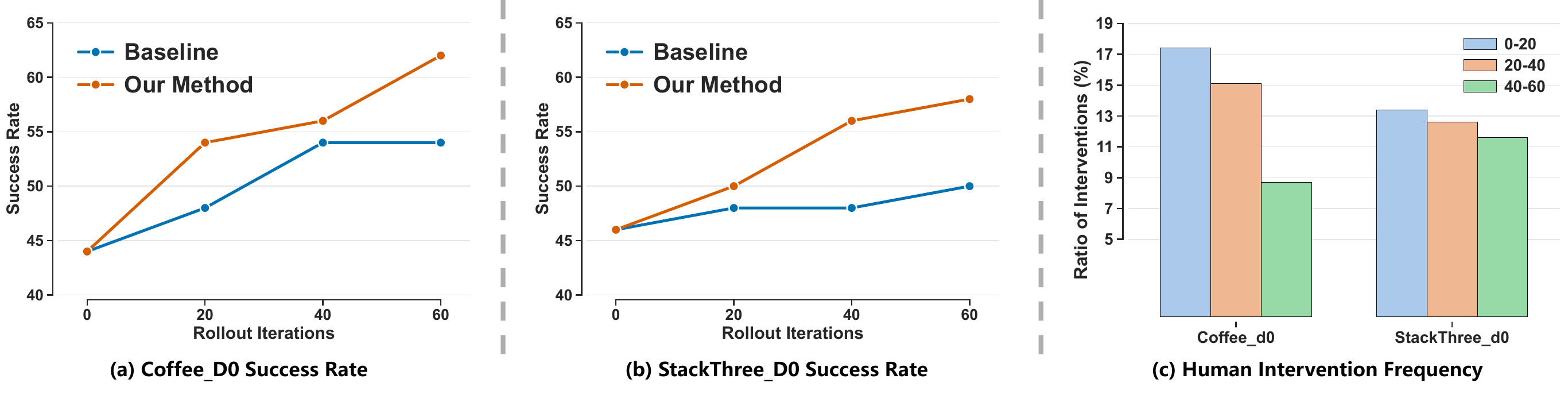}

    \caption{Lifelong learning results of APO method.
    }

    \label{fig:continual}
\end{figure*}

\subsection{Generalization to Novel Tasks}\label{exp_generlization}

In this section, we assess APO's generalization capability under three novel scenarios, as illustrated in Figure~\ref{fig:disruption}. (1) \textbf{Position Disruption}: For the Square\_D0 task, we replace the fixed initial stick position with randomized placements within a bounded operational area. (2) \textbf{Background Disruption}: In the StackThree\_D0 task, we substitute the default white background with a gray one. (3) \textbf{Texture Disruption}: In the ThreePiece\_D0 task,  the original red blocks are transmuted to wood-grain visual properties. These experiments systematically evaluate robustness against spatial, background, and visual texture variations. To fine-tune the base model on novel disruption scenarios, we collect 20 interaction trajectories under disruption scenarios and combine them with 20 expert demonstrations from the original task for subsequent fine-tuning.

Our objective is to develop an action preference optimization method that facilitates continuous improvement, enabling performance enhancements in novel disruption scenarios while retaining original task capabilities during model fine-tuning. Thus, we evaluate the performance of the fine-tuned model across both disruption scenarios and original scenarios. 

As shown in Tabel~\ref{tab:var_result}, the base policy exhibits some degree of performance degradation in disruption scenarios. However, the performance decline is relatively minor in cases of background disruption, whereas disruptions in object texture and position significantly impact performance. Both behavior cloning methods and preference optimization methods struggle to achieve significant performance improvements in novel disruption scenarios. In contrast, APO can effectively adapt to new disruption scenarios through adaptive reweighting.

Table ~\ref{tab:var_ori_results} presents the performance of the optimized model after being fine-tuned from disruption data on the original task. The results reveal that behavior cloning methods exhibit severe catastrophic forgetting, resulting in substantial performance degradation. By contrast, the preference optimization method achieves mitigated performance decline with the constraints of the reference model. 


Besides, APO utilizes adaptive reweighting to effectively integrate knowledge from both expert demonstrations and interaction trajectories. This mechanism not only facilitates learning from diverse data sources but also leads to improved performance on the original task.

\subsection{The Performance of Lifelong Learning}\label{exp_lifelong}

To investigate whether APO can iteratively improve via environment interaction, we deploy APO to interact with environments while updating the model every 20 interaction rollouts. e provide comparison results using a behavior cloning policy trained with the same number of expert demonstrations as our baseline.
For each updated model, we conduct 50 trials and report the success rate.

As shown in Figure~\ref{fig:continual}(a-b), APO achieves superior performance compared to the baseline, demonstrating its ability to effectively leverage sub-optimal human intervention trajectories for iterative model improvement. 
When the base policy exhibits diminishing improvement with increasing expert demonstrations, APO enables continual performance gains from the interaction trajectories. Besides, this improvement trend is accompanied by a corresponding reduction in the required human intervention ratio, as shown in Figure~\ref{fig:continual}(c)

\subsection{Generalization to various VLA models}\label{ref:various_vla}

To validate that APO can be adapted to different VLA models, we applied APO to fine-tune the $\pi0$-FAST~\cite{pertsch2025fast} model. $\pi0$-FAST applies discrete cosine transform encoding to encode the action chunking into discrete tokens for VLA training.  To adopt this model for downstream tasks, We regenerate the action tokenizer with 5 action chunking step for each task. 

As shown in Table~\ref{tab:pifast_result}, the base model could achieve a higher success rate compared with the OpenVLA model, benefiting from its ability to predict action chunking for robotic manipulation. Further, we compare APO with both the behavior cloning method and preference optimization method, the results demonstrate that APO could achieve consistent improvement for the $\pi0$-FAST fine-tuning. The results prove that APO could be applied to the fine-tuning of various VLA models, achieving consistent performance gains.

\begin{figure*}[t]
    \begin{minipage}[t]{0.48\textwidth} 
            \captionof{table}{The results on $\pi0$-FAST model.} 
        \centering
        \scriptsize
        \begin{tabular}{cccc}
            \toprule
            Methods   & Coffee\_D0  & StackThree\_D0 &  Insert Square \\ 
            \midrule
            Base policy & 68\% & 64\%  & 85\% \\
            Dagger & 64\% & 66\% & 85\% \\
            TPO & 48\% & 52\%  & 90\% \\
            \midrule
            APO & \textbf{76\%} & \textbf{74\%} & \textbf{95\%}\\
            \bottomrule
        \end{tabular}

        \label{tab:pifast_result}
    \end{minipage}%
    \hfill
    \begin{minipage}[t]{0.48\textwidth} 
        \captionof{table}{The results on real-world experiments.} 
        \centering
        \scriptsize
        \begin{tabular}{ccccc}
            \toprule
            Methods   & In Dis.  & Pos Dis. & Bg Dis. & Tex Dis. \\ 
            \midrule
            Base policy & 65\% & 25\% & 10\% & 25\% \\
            Dagger & 65\% & 10\% & 10\% & 25\% \\
            TPO & 75\% & 40\% & 20\% & 45\% \\
            \midrule
            APO & \textbf{85\%} & \textbf{55\%} & \textbf{30\%} & \textbf{55\%} \\
            \bottomrule
        \end{tabular}

        \label{tab:real_worlds}
    \end{minipage}
    \vspace{-1em}
\end{figure*}

\begin{figure*}[t]
    \centering
    \includegraphics[width=\linewidth]{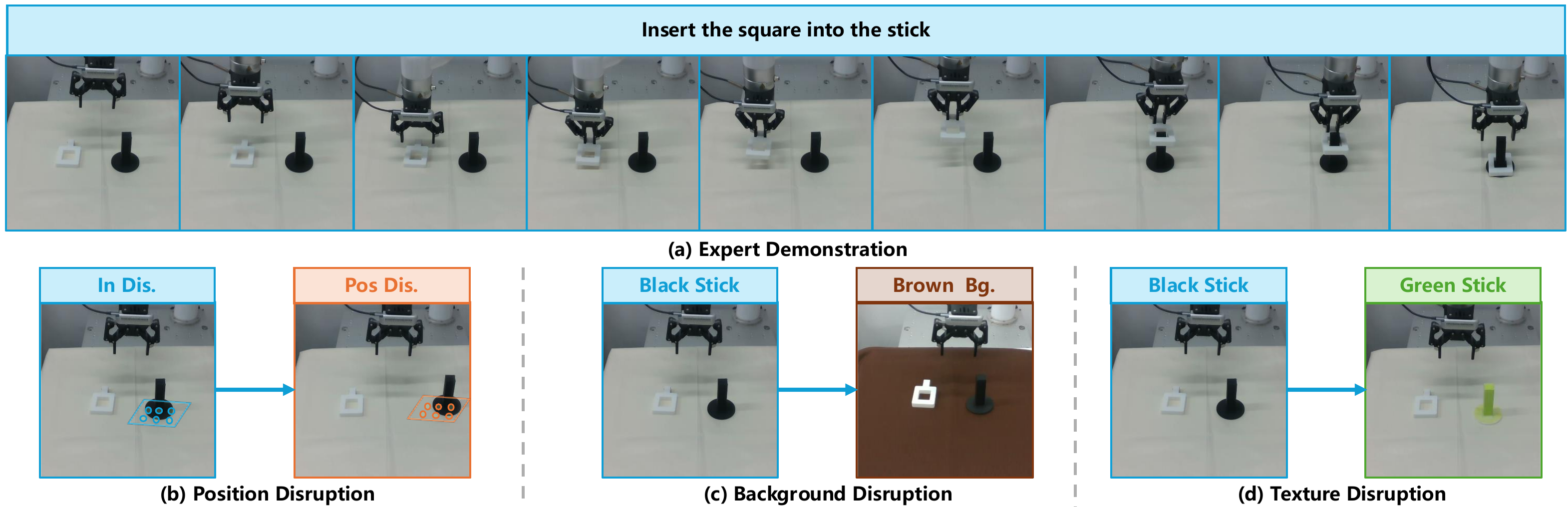}

    \caption{Demonstrations of real-world experiments with disruption settings.
    }

    \label{fig:real_world}
\end{figure*}



\subsection{Real-world Experiments}\label{exp_real}

In this work, we conduct the challenging fine-grained robotic manipulation task ``Insert the square into the stick'' as shown in Figure~\ref{fig:real_world}(a), which requires the robot to grasp the square and precisely insert into the stick. To collect expert demonstrations, we utilize the spacemouse device to gather 100 high-quality trajectories at an action frequency of 20 \textit{Hz}.  We fine-tune the OpenVLA model with the collected demonstration as the base model. Further, we deploy the base model to interact with environments and propose the real-time human-in-the-loop interventions to collect 20 interaction trajectories for subsequent action preference optimization. All methods are evaluated under the same experimental setup, and we report the average success rate from 20 trials. For a comprehensive evaluation in real-world scenarios, APO was also tested on the ``hang cup on the rack'' and ``put lemon on the plate'' tasks. 

To comprehensively evaluate APO,  experiments are conducted not only under in-distribution but also across three distinct disruption settings as shown in Figure~\ref{fig:real_world}(b-e): (1) \textbf{Position Disruption:} We change the position distribution of the stick. (2) \textbf{Background Disruption:} We replace the tablecloth from white to brown. (3) \textbf{Texture Disruption:} We replace the black stick to the green one. 

As demonstrated in Table~\ref{tab:real_worlds}, APO demonstrated robust adaptability to these downstream disruption scenarios. The results empirically validate the method’s practical utility for real-world deployment in unstructured environments.
We also adopt APO to fine-tune the $\pi_0$-FAST model in the real world scenario. The results in Table~\ref{tab:pifast_result} prove APO could achieve consistent performance gains over other methods.

\subsection{Correction from Failure scenarios}\label{analysis}

\begin{figure*}[h]
    \centering
    \includegraphics[width=\linewidth]{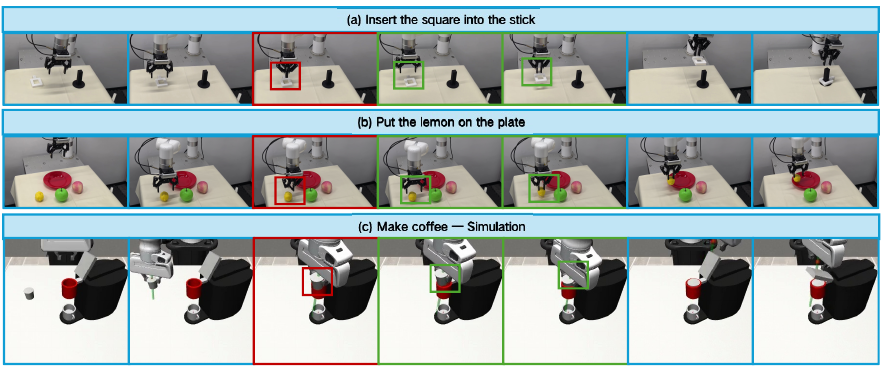}
    \caption{The rollout trajectory of APO. As indicated by the bold red and green boxes, APO can autonomously correct form failure scenarios.}
    \label{fig:correction}
\end{figure*}

In this work, we propose the APO method that enables models not only to avoid failure modes but also to self-correct within failure scenarios. As shown in Figure~\ref{fig:correction}, we provide examples of failure correction across multiple tasks, demonstrating the corrective strategies learned by APO.

For instance, in Figure~\ref{fig:correction}(a,b), when the model initially fails to grasp an object, APO identifies the failure and initiates a re-grasp attempt. Similarly, in Figure~\ref{fig:correction}(c), when a precise insertion operation is obstructed, APO learns to iteratively adjust its gripper position until the insertion is successfully completed. These examples illustrate that APO has successfully learned to recover from common failure scenarios, which directly contributes to its improved overall performance.

\section{Conclusion}
In this work, we introduce the Action Preference Optimization (APO) method to fully exploit valuable information in
failure trajectories while maintaining the stability required for large-scale VLA models. This method builds on a human-robot collaboration framework for reliable deployment, and utilizes an adaptive reweighting preference optimization algorithm with action-level binary desirability signals for stable VLA model optimization. Through APO, we could promote continuous improvement during the deployment of VLA models. We hope APO could bring insights for efficient and effective VLA model adaptation on downstream manipulation tasks.

\textbf{Discussion and Future Work.}\label{sec:limitation} While our work study the preference alignment optimization for VLA models, the experiments are based solely on autoregressive VLA models. Future work should explore a broader range of VLA frameworks, including regression-based approaches and diffusion policy models, to ensure the generalizability of our method across different architectures.

%% file: sections/appendix.tex
\section{Summary}

\begin{figure*}[h]
    \centering
    \includegraphics[width=\linewidth]{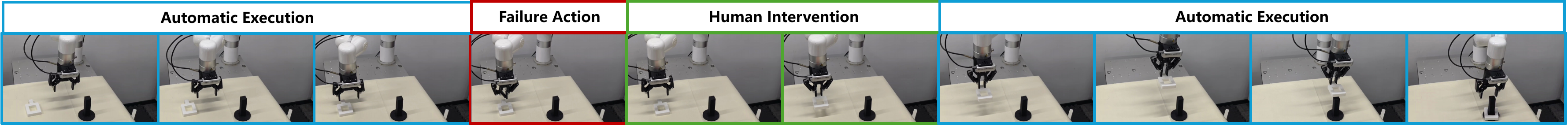}
    \caption{The demonstration of our human-assisted interaction trajectory.}\label{fig:real_world_intervention}

\end{figure*}

In this work, we propose the action preference optimization method to correct interaction failure and achieve stable optimization for VLA models. In the supplementary video, we illustrate our human-assisted interaction trajectories collection process as demonstrated in Figure~\ref{fig:real_world_intervention}. We also provide comparison videos against other methods, highlighting the effectiveness of our approach in both real-world and simulation scenarios.

\section{Human-assisted Collaboration Deployment}

In this work, we propose a human-assisted collaboration deployment framework to support reliable deployment and interaction trajectory collection. 
The \textcolor{blue}{blue block} in Figure~\ref{fig:real_world_intervention} illustrates the initial deployment of the base policy for autonomous environment interaction.
However, the base policy is trained solely on expert demonstrations. When its predicted action causes failures, this model struggles to recover from these failure states, as shown in the \textcolor{red}{red block}. To address this, we provide human intervention to manually adjust the robotic arm's movements for failure correction, as shown in the \textcolor[rgb]{0.0,0.5,0.0}{green blocks}.

Through this human-assisted approach, we ensure reliable deployment of the model in manipulation tasks. Furthermore, we annotate these interaction trajectories for subsequent preference learning. Specifically, we designate the last 10 actions before human intervention as undesirable data (representing failure actions), while the remaining trajectories serve as desirable data.

\section{Implementation Details}

In our work, we build the utility function $v$ as below to estimate the relative gain on the robotic data:
\begin{equation}
v(o, \hat{a}) = 
\begin{cases} 
\lambda_D \sigma\left(r_\theta(o, \hat{a}) - z_0  \right) & \text{if } \hat{a} \sim  \hat{a}_{\text{desirable}} \\
\lambda_U \sigma\left(  z_0 - r_\theta(o, \hat{a})  \right) & \text{if } \hat{a} \sim \hat{a}_{\text{undesirable}} ,
\end{cases}
\label{eq:v}
\end{equation}
where $z_0 = KL(\pi_{\theta}||\pi_{ref})$ to guide the model to learn from preference pair data while simultaneously preserving knowledge acquired from prior models. We compute the KL-divergence $z_0$ by leveraging the KTO~\cite{ethayarajh2024kto} method, which leverages mismatched sample pairs for KL estimation. Further, we ignore the reject reward of the gripper action token to prevent erroneous rejection of the same gripper state.

\begin{figure*}[t]
    \centering
    \includegraphics[width=\linewidth]{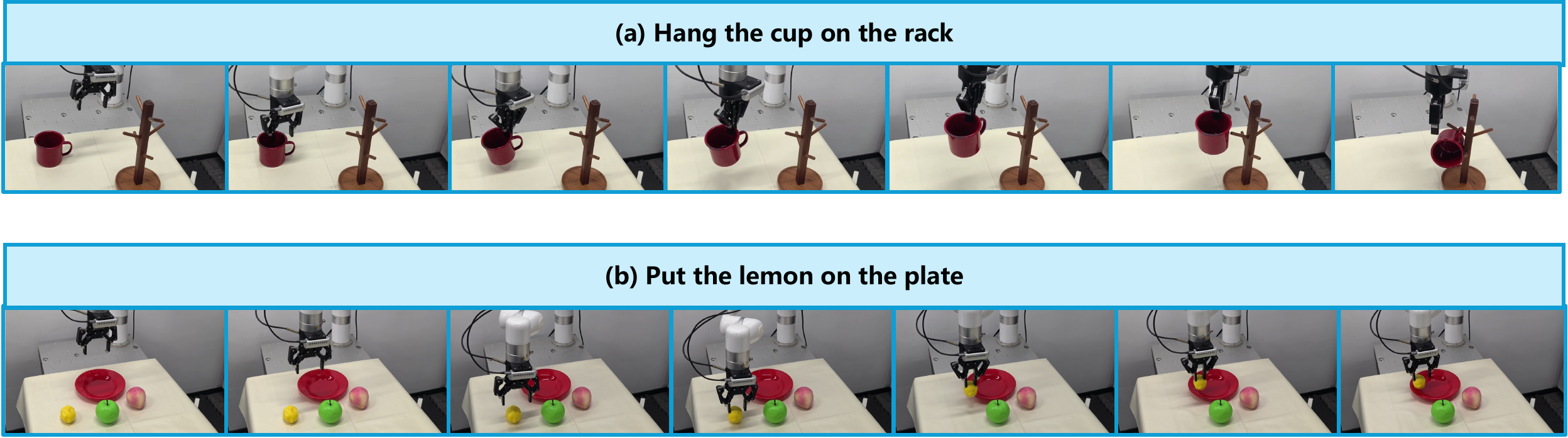}
    \caption{The demonstrations of real-world experiments.}
    \label{fig:real_world}
\end{figure*}

\begin{figure*}[thb]
    \begin{minipage}[t]{0.48\textwidth} 
            \captionof{table}{The results on $\pi0$-FAST model.} 
            \label{tab:pi}
        \centering
        \begin{tabular}{ccc}
            \toprule
            Methods   & Square  \\ 
            \midrule
            Base policy & 85\%  \\
            Dagger & 85\%  \\
            TPO & 90\% \\
            \midrule
            Ours & \textbf{95\%} \\
            \bottomrule
        \end{tabular}

        \label{tab:pifast_result}
    \end{minipage}%
    \hfill
    \begin{minipage}[t]{0.48\textwidth} 
        \captionof{table}{The results on real-world experiments.} 
        \label{tab:real}
        \centering
        \begin{tabular}{ccccc}
            \toprule
            Methods   & Hang  & Put\\ 
            \midrule
            Base policy & 70\% & 85\% \\
            Dagger & 65\% & 85\% \\
            TPO & 75\% & 80\% \\
            \midrule
            Ours & \textbf{90\%} & \textbf{100\%} \\
            \bottomrule
        \end{tabular}
    \end{minipage}
    \vspace{-1em}
\end{figure*}

\section{More Real-world Experiments}

\subsection{Generalization to various VLA models}
In this section, we adopt our method to fine-tune the $\pi_0$-FAST model. As shown in Table~\ref{tab:pi}, the $\pi_0$-FAST model achieves a higher success rate, benefiting from its action chunking prediction. Besides, our method could achieve consistent performance gains in real-world experiments. Because our method needs to decode to continuous action for adaptive reweighting, however, the $\pi_0$-FAST model may fail to decode predicted action tokens into meaningful continuous actions, thus when the predicted action token sequences cannot be decoded to x, we would set the weight as 1 to promote the model focus on predicting correct action token sequences.

\subsection{More real-world tasks}
In this section, we provide two more real-world experiments as shown in Figure~\ref{fig:real_world}. For each task, we collect 100 expert demonstrations to train the base policy. Further, we deploy the base policy to interact with
environments and collect 20 human-intervened trajectories. We mix the 20 human-intervened trajectories with 20 expert demonstrations for model preference optimization. As shown in Table~\ref{tab:real}, our method could achieve better performance compared with other behavior cloning and preference optimization methods.

%% file: paper.bbl
\begin{thebibliography}{50}
\providecommand{\natexlab}[1]{#1}
\providecommand{\url}[1]{\texttt{#1}}
\expandafter\ifx\csname urlstyle\endcsname\relax
  \providecommand{\doi}[1]{doi: #1}\else
  \providecommand{\doi}{doi: \begingroup \urlstyle{rm}\Url}\fi

\bibitem[Argall et~al.(2009)Argall, Chernova, Veloso, and Browning]{survey09}
Brenna~D. Argall, Sonia Chernova, Manuela Veloso, and Brett Browning.
\newblock A survey of robot learning from demonstration.
\newblock \emph{Robotics and Autonomous Systems}, 57\penalty0 (5):\penalty0 469--483, 2009.
\newblock ISSN 0921-8890.
\newblock \doi{https://doi.org/10.1016/j.robot.2008.10.024}.
\newblock URL \url{https://www.sciencedirect.com/science/article/pii/S0921889008001772}.

\bibitem[Bai et~al.(2022)Bai, Jones, Ndousse, Askell, Chen, DasSarma, Drain, Fort, Ganguli, Henighan, et~al.]{bai2022training}
Yuntao Bai, Andy Jones, Kamal Ndousse, Amanda Askell, Anna Chen, Nova DasSarma, Dawn Drain, Stanislav Fort, Deep Ganguli, Tom Henighan, et~al.
\newblock Training a helpful and harmless assistant with reinforcement learning from human feedback.
\newblock \emph{arXiv preprint arXiv:2204.05862}, 2022.

\bibitem[Bjorck et~al.(2025)Bjorck, Casta{\~n}eda, Cherniadev, Da, Ding, Fan, Fang, Fox, Hu, Huang, et~al.]{bjorck2025gr00t}
Johan Bjorck, Fernando Casta{\~n}eda, Nikita Cherniadev, Xingye Da, Runyu Ding, Linxi Fan, Yu~Fang, Dieter Fox, Fengyuan Hu, Spencer Huang, et~al.
\newblock Gr00t n1: An open foundation model for generalist humanoid robots.
\newblock \emph{arXiv preprint arXiv:2503.14734}, 2025.

\bibitem[Black et~al.(2024)Black, Brown, Driess, Esmail, Equi, Finn, Fusai, Groom, Hausman, Ichter, et~al.]{black2024pi0}
Kevin Black, Noah Brown, Danny Driess, Adnan Esmail, Michael Equi, Chelsea Finn, Niccolo Fusai, Lachy Groom, Karol Hausman, Brian Ichter, et~al.
\newblock pi0 : A vision-language-action flow model for general robot control.
\newblock \emph{arXiv preprint arXiv:2410.24164}, 2024.

\bibitem[Brohan et~al.(2023)Brohan, Brown, Carbajal, Chebotar, Chen, Choromanski, Ding, Driess, Dubey, Finn, et~al.]{brohan2023rt}
Anthony Brohan, Noah Brown, Justice Carbajal, Yevgen Chebotar, Xi~Chen, Krzysztof Choromanski, Tianli Ding, Danny Driess, Avinava Dubey, Chelsea Finn, et~al.
\newblock Rt-2: Vision-language-action models transfer web knowledge to robotic control.
\newblock \emph{arXiv preprint arXiv:2307.15818}, 2023.

\bibitem[Bu et~al.(2025)Bu, Cai, Chen, Cui, Ding, Feng, Gao, He, Huang, Jiang, et~al.]{bu2025agibot}
Qingwen Bu, Jisong Cai, Li~Chen, Xiuqi Cui, Yan Ding, Siyuan Feng, Shenyuan Gao, Xindong He, Xu~Huang, Shu Jiang, et~al.
\newblock Agibot world colosseo: A large-scale manipulation platform for scalable and intelligent embodied systems.
\newblock \emph{arXiv preprint arXiv:2503.06669}, 2025.

\bibitem[Celemin et~al.(2022)Celemin, P{\'e}rez-Dattari, Chisari, Franzese, de~Souza~Rosa, Prakash, Ajanovi{\'c}, Ferraz, Valada, Kober, et~al.]{celemin2022interactive}
Carlos Celemin, Rodrigo P{\'e}rez-Dattari, Eugenio Chisari, Giovanni Franzese, Leandro de~Souza~Rosa, Ravi Prakash, Zlatan Ajanovi{\'c}, Marta Ferraz, Abhinav Valada, Jens Kober, et~al.
\newblock Interactive imitation learning in robotics: A survey.
\newblock \emph{Foundations and Trends{\textregistered} in Robotics}, 10\penalty0 (1-2):\penalty0 1--197, 2022.

\bibitem[Cheang et~al.(2024)Cheang, Chen, Jing, Kong, Li, Li, Liu, Wu, Xu, Yang, et~al.]{cheang2024gr}
Chi-Lam Cheang, Guangzeng Chen, Ya~Jing, Tao Kong, Hang Li, Yifeng Li, Yuxiao Liu, Hongtao Wu, Jiafeng Xu, Yichu Yang, et~al.
\newblock Gr-2: A generative video-language-action model with web-scale knowledge for robot manipulation.
\newblock \emph{arXiv preprint arXiv:2410.06158}, 2024.

\bibitem[Christiano et~al.(2017)Christiano, Leike, Brown, Martic, Legg, and Amodei]{christiano2017deep}
Paul~F Christiano, Jan Leike, Tom Brown, Miljan Martic, Shane Legg, and Dario Amodei.
\newblock Deep reinforcement learning from human preferences.
\newblock \emph{Advances in neural information processing systems}, 30, 2017.

\bibitem[Cui et~al.(2023)Cui, Karamcheti, Palleti, Shivakumar, Liang, and Sadigh]{cui2023no}
Yuchen Cui, Siddharth Karamcheti, Raj Palleti, Nidhya Shivakumar, Percy Liang, and Dorsa Sadigh.
\newblock No, to the right: Online language corrections for robotic manipulation via shared autonomy.
\newblock In \emph{Proceedings of the 2023 ACM/IEEE International Conference on Human-Robot Interaction}, HRI '23, page 93–101, New York, NY, USA, 2023. Association for Computing Machinery.
\newblock ISBN 9781450399647.
\newblock \doi{10.1145/3568162.3578623}.
\newblock URL \url{https://doi.org/10.1145/3568162.3578623}.

\bibitem[Ethayarajh et~al.(2024)Ethayarajh, Xu, Muennighoff, Jurafsky, and Kiela]{ethayarajh2024kto}
Kawin Ethayarajh, Winnie Xu, Niklas Muennighoff, Dan Jurafsky, and Douwe Kiela.
\newblock Kto: Model alignment as prospect theoretic optimization.
\newblock \emph{arXiv preprint arXiv:2402.01306}, 2024.

\bibitem[Fang et~al.(2023)Fang, Fang, Tang, Liu, Wang, Wang, Zhu, and Lu]{fang2023rh20t}
Hao-Shu Fang, Hongjie Fang, Zhenyu Tang, Jirong Liu, Chenxi Wang, Junbo Wang, Haoyi Zhu, and Cewu Lu.
\newblock Rh20t: A comprehensive robotic dataset for learning diverse skills in one-shot.
\newblock \emph{arXiv preprint arXiv:2307.00595}, 2023.

\bibitem[Gao et~al.(2023)Gao, Schulman, and Hilton]{gao2023scaling}
Leo Gao, John Schulman, and Jacob Hilton.
\newblock Scaling laws for reward model overoptimization.
\newblock In \emph{International Conference on Machine Learning}, pages 10835--10866. PMLR, 2023.

\bibitem[Gokmen et~al.(2023)Gokmen, Ho, and Khansari]{gokmen2023asking}
Cem Gokmen, Daniel Ho, and Mohi Khansari.
\newblock Asking for help: Failure prediction in behavioral cloning through value approximation.
\newblock In \emph{2023 IEEE International Conference on Robotics and Automation (ICRA)}, pages 5821--5828. IEEE, 2023.

\bibitem[Hoque et~al.(2023)Hoque, Chen, Sharma, Dharmarajan, Thananjeyan, Abbeel, and Goldberg]{hoque2023fleet}
Ryan Hoque, Lawrence~Yunliang Chen, Satvik Sharma, Karthik Dharmarajan, Brijen Thananjeyan, Pieter Abbeel, and Ken Goldberg.
\newblock Fleet-dagger: Interactive robot fleet learning with scalable human supervision.
\newblock In \emph{Conference on Robot Learning}, pages 368--380. PMLR, 2023.

\bibitem[Hu et~al.(2022)Hu, Shen, Wallis, Allen-Zhu, Li, Wang, Wang, Chen, et~al.]{hu2022lora}
Edward~J Hu, Yelong Shen, Phillip Wallis, Zeyuan Allen-Zhu, Yuanzhi Li, Shean Wang, Lu~Wang, Weizhu Chen, et~al.
\newblock Lora: Low-rank adaptation of large language models.
\newblock \emph{ICLR}, 1\penalty0 (2):\penalty0 3, 2022.

\bibitem[Kelly et~al.(2019)Kelly, Sidrane, Driggs-Campbell, and Kochenderfer]{kelly2019hg}
Michael Kelly, Chelsea Sidrane, Katherine Driggs-Campbell, and Mykel~J Kochenderfer.
\newblock Hg-dagger: Interactive imitation learning with human experts.
\newblock In \emph{2019 International Conference on Robotics and Automation (ICRA)}, pages 8077--8083. IEEE, 2019.

\bibitem[Kim et~al.(2024)Kim, Pertsch, Karamcheti, Xiao, Balakrishna, Nair, Rafailov, Foster, Lam, Sanketi, et~al.]{kim2024openvla}
Moo~Jin Kim, Karl Pertsch, Siddharth Karamcheti, Ted Xiao, Ashwin Balakrishna, Suraj Nair, Rafael Rafailov, Ethan Foster, Grace Lam, Pannag Sanketi, et~al.
\newblock Openvla: An open-source vision-language-action model.
\newblock \emph{arXiv preprint arXiv:2406.09246}, 2024.

\bibitem[Li et~al.(2024)Li, Zhang, Guo, Zhang, Li, Zhang, Zhang, Zhang, Li, Liu, et~al.]{li2024llava}
Bo~Li, Yuanhan Zhang, Dong Guo, Renrui Zhang, Feng Li, Hao Zhang, Kaichen Zhang, Peiyuan Zhang, Yanwei Li, Ziwei Liu, et~al.
\newblock Llava-onevision: Easy visual task transfer.
\newblock \emph{arXiv preprint arXiv:2408.03326}, 2024.

\bibitem[Li et~al.(2022)Li, Peng, and Zhou]{li2022efficient}
Quanyi Li, Zhenghao Peng, and Bolei Zhou.
\newblock Efficient learning of safe driving policy via human-ai copilot optimization.
\newblock \emph{arXiv preprint arXiv:2202.10341}, 2022.

\bibitem[Ling-Huan et~al.(2023)Ling-Huan, Wei, Wen-Shi, Hui, and Yao-Nan]{MIRwork}
Kong Ling-Huan, He~Wei, Chen Wen-Shi, Zhang Hui, and Wang Yao-Nan.
\newblock Dynamic movement primitives based robot skills learning.
\newblock \emph{Machine Intelligence Research}, 20:\penalty0 396--407, 2023.
\newblock \doi{10.1007/s11633-022-1346-z}.

\bibitem[Liu et~al.(2023)Liu, Nasiriany, Zhang, Bao, and Zhu]{liu2022robot}
Huihan Liu, Soroush Nasiriany, Lance Zhang, Zhiyao Bao, and Yuke Zhu.
\newblock Robot learning on the job: Human-in-the-loop autonomy and learning during deployment.
\newblock In \emph{Robotics: Science and Systems (RSS)}, 2023.

\bibitem[Liu et~al.(2024{\natexlab{a}})Liu, Dass, Mart{\'\i}n-Mart{\'\i}n, and Zhu]{liu2024model}
Huihan Liu, Shivin Dass, Roberto Mart{\'\i}n-Mart{\'\i}n, and Yuke Zhu.
\newblock Model-based runtime monitoring with interactive imitation learning.
\newblock In \emph{2024 IEEE International Conference on Robotics and Automation (ICRA)}, pages 4154--4161. IEEE, 2024{\natexlab{a}}.

\bibitem[Liu et~al.(2024{\natexlab{b}})Liu, Wu, Li, Tan, Chen, Wang, Xu, Su, and Zhu]{liu2024rdt}
Songming Liu, Lingxuan Wu, Bangguo Li, Hengkai Tan, Huayu Chen, Zhengyi Wang, Ke~Xu, Hang Su, and Jun Zhu.
\newblock Rdt-1b: a diffusion foundation model for bimanual manipulation.
\newblock \emph{arXiv preprint arXiv:2410.07864}, 2024{\natexlab{b}}.

\bibitem[Lu et~al.(2024)Lu, Xia, Wang, Wang, Zhao, Hu, and Li]{lu2024koi}
Jingxian Lu, Wenke Xia, Dong Wang, Zhigang Wang, Bin Zhao, Di~Hu, and Xuelong Li.
\newblock Koi: Accelerating online imitation learning via hybrid key-state guidance.
\newblock \emph{arXiv preprint arXiv:2408.02912}, 2024.

\bibitem[Luo et~al.(2023)Luo, Dong, Zhai, Ma, and Levine]{luo2023rlif}
Jianlan Luo, Perry Dong, Yuexiang Zhai, Yi~Ma, and Sergey Levine.
\newblock Rlif: Interactive imitation learning as reinforcement learning.
\newblock \emph{arXiv preprint arXiv:2311.12996}, 2023.

\bibitem[Luo et~al.(2024)Luo, Xu, Wu, and Levine]{luo2024precise}
Jianlan Luo, Charles Xu, Jeffrey Wu, and Sergey Levine.
\newblock Precise and dexterous robotic manipulation via human-in-the-loop reinforcement learning.
\newblock \emph{arXiv preprint arXiv:2410.21845}, 2024.

\bibitem[Mandlekar et~al.(2020)Mandlekar, Xu, Mart{\'\i}n-Mart{\'\i}n, Zhu, Fei-Fei, and Savarese]{mandlekar2020human}
Ajay Mandlekar, Danfei Xu, Roberto Mart{\'\i}n-Mart{\'\i}n, Yuke Zhu, Li~Fei-Fei, and Silvio Savarese.
\newblock Human-in-the-loop imitation learning using remote teleoperation.
\newblock \emph{arXiv preprint arXiv:2012.06733}, 2020.

\bibitem[Mandlekar et~al.(2021)Mandlekar, Xu, Wong, Nasiriany, Wang, Kulkarni, Fei-Fei, Savarese, Zhu, and Mart\'{i}n-Mart\'{i}n]{robomimic2021}
Ajay Mandlekar, Danfei Xu, Josiah Wong, Soroush Nasiriany, Chen Wang, Rohun Kulkarni, Li~Fei-Fei, Silvio Savarese, Yuke Zhu, and Roberto Mart\'{i}n-Mart\'{i}n.
\newblock What matters in learning from offline human demonstrations for robot manipulation.
\newblock In \emph{arXiv preprint arXiv:2108.03298}, 2021.

\bibitem[Nakano et~al.(2021)Nakano, Hilton, Balaji, Wu, Ouyang, Kim, Hesse, Jain, Kosaraju, Saunders, et~al.]{nakano2021webgpt}
Reiichiro Nakano, Jacob Hilton, Suchir Balaji, Jeff Wu, Long Ouyang, Christina Kim, Christopher Hesse, Shantanu Jain, Vineet Kosaraju, William Saunders, et~al.
\newblock Webgpt: Browser-assisted question-answering with human feedback.
\newblock \emph{arXiv preprint arXiv:2112.09332}, 2021.

\bibitem[Ouyang et~al.(2022)Ouyang, Wu, Jiang, Almeida, Wainwright, Mishkin, Zhang, Agarwal, Slama, Ray, et~al.]{ouyang2022training}
Long Ouyang, Jeffrey Wu, Xu~Jiang, Diogo Almeida, Carroll Wainwright, Pamela Mishkin, Chong Zhang, Sandhini Agarwal, Katarina Slama, Alex Ray, et~al.
\newblock Training language models to follow instructions with human feedback.
\newblock \emph{Advances in neural information processing systems}, 35:\penalty0 27730--27744, 2022.

\bibitem[Pang et~al.(2024)Pang, Xia, Wang, Zhao, Hu, Wang, and Li]{pang2024}
Xincheng Pang, Wenke Xia, Zhigang Wang, Bin Zhao, Di~Hu, Dong Wang, and Xuelong Li.
\newblock Depth helps: Improving pre-trained rgb-based policy with depth information injection.
\newblock In \emph{2024 IEEE/RSJ International Conference on Intelligent Robots and Systems (IROS)}, pages 7251--7256, Oct 2024.
\newblock \doi{10.1109/IROS58592.2024.10802706}.

\bibitem[Pertsch et~al.(2025)Pertsch, Stachowicz, Ichter, Driess, Nair, Vuong, Mees, Finn, and Levine]{pertsch2025fast}
Karl Pertsch, Kyle Stachowicz, Brian Ichter, Danny Driess, Suraj Nair, Quan Vuong, Oier Mees, Chelsea Finn, and Sergey Levine.
\newblock Fast: Efficient action tokenization for vision-language-action models.
\newblock \emph{arXiv preprint arXiv:2501.09747}, 2025.

\bibitem[Rafailov et~al.(2023)Rafailov, Sharma, Mitchell, Manning, Ermon, and Finn]{rafailov2023direct}
Rafael Rafailov, Archit Sharma, Eric Mitchell, Christopher~D Manning, Stefano Ermon, and Chelsea Finn.
\newblock Direct preference optimization: Your language model is secretly a reward model.
\newblock \emph{Advances in Neural Information Processing Systems}, 36:\penalty0 53728--53741, 2023.

\bibitem[Ross et~al.(2011)Ross, Gordon, and Bagnell]{ross2011reduction}
St{\'e}phane Ross, Geoffrey Gordon, and Drew Bagnell.
\newblock A reduction of imitation learning and structured prediction to no-regret online learning.
\newblock In \emph{Proceedings of the fourteenth international conference on artificial intelligence and statistics}, pages 627--635. JMLR Workshop and Conference Proceedings, 2011.

\bibitem[Schulman et~al.(2017)Schulman, Wolski, Dhariwal, Radford, and Klimov]{schulman2017proximal}
John Schulman, Filip Wolski, Prafulla Dhariwal, Alec Radford, and Oleg Klimov.
\newblock Proximal policy optimization algorithms.
\newblock \emph{arXiv preprint arXiv:1707.06347}, 2017.

\bibitem[Stiennon et~al.(2020)Stiennon, Ouyang, Wu, Ziegler, Lowe, Voss, Radford, Amodei, and Christiano]{stiennon2020learning}
Nisan Stiennon, Long Ouyang, Jeffrey Wu, Daniel Ziegler, Ryan Lowe, Chelsea Voss, Alec Radford, Dario Amodei, and Paul~F Christiano.
\newblock Learning to summarize with human feedback.
\newblock \emph{Advances in neural information processing systems}, 33:\penalty0 3008--3021, 2020.

\bibitem[Team et~al.(2024)Team, Ghosh, Walke, Pertsch, Black, Mees, Dasari, Hejna, Kreiman, Xu, et~al.]{team2024octo}
Octo~Model Team, Dibya Ghosh, Homer Walke, Karl Pertsch, Kevin Black, Oier Mees, Sudeep Dasari, Joey Hejna, Tobias Kreiman, Charles Xu, et~al.
\newblock Octo: An open-source generalist robot policy.
\newblock \emph{arXiv preprint arXiv:2405.12213}, 2024.

\bibitem[Touvron et~al.(2023)Touvron, Lavril, Izacard, Martinet, Lachaux, Lacroix, Rozi{\`e}re, Goyal, Hambro, Azhar, et~al.]{touvron2023llama}
Hugo Touvron, Thibaut Lavril, Gautier Izacard, Xavier Martinet, Marie-Anne Lachaux, Timoth{\'e}e Lacroix, Baptiste Rozi{\`e}re, Naman Goyal, Eric Hambro, Faisal Azhar, et~al.
\newblock Llama: Open and efficient foundation language models.
\newblock \emph{arXiv preprint arXiv:2302.13971}, 2023.

\bibitem[Tversky and Kahneman(1992)]{tversky1992advances}
Amos Tversky and Daniel Kahneman.
\newblock Advances in prospect theory: Cumulative representation of uncertainty.
\newblock \emph{Journal of Risk and uncertainty}, 5:\penalty0 297--323, 1992.

\bibitem[Vuong et~al.(2023)Vuong, Levine, Walke, Pertsch, Singh, Doshi, Xu, Luo, Tan, Shah, et~al.]{vuong2023open}
Quan Vuong, Sergey Levine, Homer~Rich Walke, Karl Pertsch, Anikait Singh, Ria Doshi, Charles Xu, Jianlan Luo, Liam Tan, Dhruv Shah, et~al.
\newblock Open x-embodiment: Robotic learning datasets and rt-x models.
\newblock In \emph{Towards Generalist Robots: Learning Paradigms for Scalable Skill Acquisition@ CoRL2023}, 2023.

\bibitem[Wei et~al.(2022)Wei, Wang, Schuurmans, Bosma, Xia, Chi, Le, Zhou, et~al.]{wei2022chain}
Jason Wei, Xuezhi Wang, Dale Schuurmans, Maarten Bosma, Fei Xia, Ed~Chi, Quoc~V Le, Denny Zhou, et~al.
\newblock Chain-of-thought prompting elicits reasoning in large language models.
\newblock \emph{Advances in neural information processing systems}, 35:\penalty0 24824--24837, 2022.

\bibitem[Wen et~al.(2025)Wen, Zhu, Li, Tang, Shen, and Feng]{wen2025dexvla}
Junjie Wen, Yichen Zhu, Jinming Li, Zhibin Tang, Chaomin Shen, and Feifei Feng.
\newblock Dexvla: Vision-language model with plug-in diffusion expert for general robot control.
\newblock \emph{arXiv preprint arXiv:2502.05855}, 2025.

\bibitem[Xia et~al.(2024)Xia, Wang, Pang, Wang, Zhao, Hu, and Li]{xia2024kinematic}
Wenke Xia, Dong Wang, Xincheng Pang, Zhigang Wang, Bin Zhao, Di~Hu, and Xuelong Li.
\newblock Kinematic-aware prompting for generalizable articulated object manipulation with llms.
\newblock In \emph{2024 IEEE International Conference on Robotics and Automation (ICRA)}, pages 2073--2080, 2024.
\newblock \doi{10.1109/ICRA57147.2024.10610744}.

\bibitem[Xia et~al.(2025)Xia, Feng, Wang, and Hu]{Xia_2025_CVPR}
Wenke Xia, Ruoxuan Feng, Dong Wang, and Di~Hu.
\newblock Phoenix: A motion-based self-reflection framework for fine-grained robotic action correction.
\newblock In \emph{Proceedings of the Computer Vision and Pattern Recognition Conference (CVPR)}, pages 6981--6990, June 2025.

\bibitem[Yel and Bezzo(2019)]{yel2019fast}
Esen Yel and Nicola Bezzo.
\newblock Fast run-time monitoring, replanning, and recovery for safe autonomous system operations.
\newblock In \emph{2019 IEEE/RSJ International Conference on Intelligent Robots and Systems (IROS)}, pages 1661--1667. IEEE, 2019.

\bibitem[Ying et~al.(2025)Ying, Lei, Yuejiao, Yi, Yi, Sicheng, Yiyi, and Lap-Pui]{MIRsurvey}
Zheng Ying, Yao Lei, Su~Yuejiao, Zhang Yi, Wang Yi, Zhao Sicheng, Zhang Yiyi, and Chau Lap-Pui.
\newblock A survey of embodied learning for object-centric robotic manipulation.
\newblock \emph{Machine Intelligence Research}, 22:\penalty0 588--626, 2025.
\newblock \doi{10.1007/s11633-025-1542-8}.

\bibitem[Zeng et~al.(2024)Zeng, Bu, Wang, Xia, Chen, Dong, Song, Wang, Hu, Luo, et~al.]{zeng2024learning}
Jia Zeng, Qingwen Bu, Bangjun Wang, Wenke Xia, Li~Chen, Hao Dong, Haoming Song, Dong Wang, Di~Hu, Ping Luo, et~al.
\newblock Learning manipulation by predicting interaction.
\newblock \emph{arXiv preprint arXiv:2406.00439}, 2024.

\bibitem[Zhang et~al.(2024)Zhang, Zheng, Chen, Jang, Li, Han, Wang, Ding, Fox, and Yao]{zhang2024grape}
Zijian Zhang, Kaiyuan Zheng, Zhaorun Chen, Joel Jang, Yi~Li, Siwei Han, Chaoqi Wang, Mingyu Ding, Dieter Fox, and Huaxiu Yao.
\newblock Grape: Generalizing robot policy via preference alignment.
\newblock \emph{arXiv preprint arXiv:2411.19309}, 2024.

\bibitem[Ziegler et~al.(2019)Ziegler, Stiennon, Wu, Brown, Radford, Amodei, Christiano, and Irving]{ziegler2019fine}
Daniel~M Ziegler, Nisan Stiennon, Jeffrey Wu, Tom~B Brown, Alec Radford, Dario Amodei, Paul Christiano, and Geoffrey Irving.
\newblock Fine-tuning language models from human preferences.
\newblock \emph{arXiv preprint arXiv:1909.08593}, 2019.

\end{thebibliography}
